# Producing a Standard Dataset of Speed Climbing Training Videos Using Deep Learning Techniques

Yufei Xie[1,2], Shaoman Li[2], Penghui Lin[2]
[1]College of Computing and Information Technologies, National University, Manila, 999005, Philippines
[2]Rock Climbing Sports Centre, Jiangxi College of Applied Technology, Ganzhou, 341000, China

**Abstract:** This dissertation presents a methodology for recording speed climbing training sessions with multiple cameras and annotating the videos with relevant data, including body position, hand and foot placement, and timing. The annotated data is then analyzed using deep learning techniques to create a standard dataset of speed climbing training videos. The results demonstrate the potential of the new dataset for improving speed climbing training and research, including identifying areas for improvement, creating personalized training plans, and analyzing the effects of different training methods. The findings will also be applied to the training process of the Jiangxi climbing team through further empirical research to test the findings and further explore the feasibility of this study.

## 1. Introduction

Speed climbing is a discipline of sport climbing that involves ascending a standardized climbing route as quickly as possible. The route is typically 15 meters tall, with pre-determined hand and foot holds that are the same for all climbers. Climbers race against each other in a knockout-style format, with the fastest climber advancing to the next round.

Speed climbing is a highly specialized and physically demanding sport that requires a unique set of skills and techniques to excel. These include explosive power, strength, agility, and precision, as well as the ability to read the route and anticipate the next move. In addition to being a standalone sport, speed climbing is also an important component of training for other disciplines of climbing, such as bouldering and lead climbing.

In recent years, speed climbing has gained increased attention and recognition, particularly with its inclusion in the Olympics for the first time in 2021. As a result, there is growing interest in improving training methods for speed climbing and enhancing the performance of athletes. Standardized datasets of speed climbing training videos, produced using deep learning techniques as proposed in this research, have the potential to revolutionize the way athletes train and compete in the sport.

## 2. Literature Review

### 2.1. Overview of the current state of speed climbing training and its limitations

The current state of speed climbing training often relies on subjective observations and limited data, leading to a lack of standardization and potential inefficiencies. Training methods typically involve climbing the speed route repeatedly, with coaches providing feedback on technique and timing. While this approach can be effective to a certain extent, it has several limitations.

Relying on subjective observations can lead to inconsistencies and biases, as different coaches may have different interpretations of what constitutes good technique and timing. This can also result in limited feedback and guidance for athletes, as coaches may not have a comprehensive understanding of the various factors that contribute to successful speed climbing.

There is a lack of standardization in training methods and data, making it difficult to compare and analyze training techniques across different athletes and coaches. This also limits the ability to identify areas for improvement and optimize training methods for individual athletes.

The limited data available on speed climbing training can also hinder research efforts to understand the sport and improve performance. Without comprehensive and standardized data, it can be difficult to identify trends and patterns in speed climbing technique and training methods.

The current state of speed climbing training has several limitations that can hinder athlete performance and research efforts. The development of a standardized dataset of speed climbing training videos using deep learning techniques, as proposed in this research, has the potential to address these limitations and improve training methods for speed climbing athletes.

## 2.2. Summary of existing datasets of climbing videos and their limitations

The purpose of this research is to produce a standard dataset of speed climbing training videos using deep learning techniques. Speed climbing is a highly specialized sport that requires a unique set of skills and techniques to excel. However, current training methods often rely on subjective observations and limited data, leading to a lack of standardization and potential inefficiencies. Deep learning techniques offer a promising avenue for creating standardized datasets and improving training methods in speed climbing and other sports.

There are few publicly available datasets of speed climbing videos, such as SPEED21 [1]. which provides 362 performances of 55 world elite athletes annotated with 2D skeleton sequences, and Schmidt et al [2]. which provides 100 performances of 20 athletes annotated with joint angles and velocities.

Some of the limitations of existing datasets include the lack of diversity in routes, wall angles, camera views and climber profiles, the difficulty in obtaining high-quality video data from competition events [3], and the lack of ground truth labels for performance metrics such as power output, energy expenditure and efficiency.

Some possible applications of speed climbing video analysis include improving motion understanding and recognition, developing coaching and feedback systems, enhancing route setting and evaluation, and creating realistic VR simulations.

## 3. Methodology

### 3.1. Recording of speed climbing training sessions with multiple cameras

Recording speed climbing training sessions with multiple cameras is a critical step in creating a standardized dataset of training videos. Multiple camera angles can provide a more comprehensive view of the climbing route and the athlete's technique, allowing for more accurate annotation and analysis.

Ideally, a minimum of three cameras should be used to record a speed climbing training session: one at the start of the route, one at the top of the route, and one positioned to capture a side view of the athlete as they climb. Additional cameras can be added as necessary to capture specific angles or movements.

To ensure consistency across recordings, the cameras should be placed in the same position for each training session. The camera settings, such as exposure and focus, should also be consistent to ensure that the video quality is standardized.

During the recording, it is essential to ensure that the climbing route and holds are visible and clearly defined in the videos. Lighting conditions should also be consistent across all recordings to avoid any discrepancies in the video quality.

Recording speed climbing training sessions with multiple cameras can provide valuable data for deep learning analysis and the creation of a standardized dataset. It can help to identify areas for improvement in technique and timing, and provide coaches with more accurate feedback for training optimization. Overall, recording speed climbing training sessions with multiple cameras is a crucial step in improving training methods and advancing the sport of speed climbing.

### 3.2. Annotation of videos with relevant data, including body position, hand and foot placement, and timing

Annotation of speed climbing training videos with relevant data is a critical step in creating a standardized dataset. The annotated data can be used to analyze the athlete's technique and timing, identify areas for improvement, and provide personalized training recommendations. The relevant data that should be annotated includes:

Body position: This refers to the position of the athlete's body during the climb, including the angle of the torso, the position of the hips and legs, and the alignment of the head and shoulders.

Hand and foot placement: This refers to the placement of the athlete's hands and feet on the holds, including the type of hold (e.g., crimp, jug, slope), the grip (e.g., open-handed, closed-handed), and the exact location of the hold on the route.

Timing: This refers to the duration of each movement, such as the time it takes for the athlete to grab a hold, shift their weight, or complete a specific move.

Annotation can be done manually or using computer vision algorithms. Manual annotation involves watching the video and recording the relevant data in a spreadsheet or other database. Computer vision algorithms can automate this process by detecting key points on the athlete's body and tracking their movement throughout the climb.

### 3.3. Use of deep learning techniques to analyze the annotated data and create a standard dataset of speed climbing training videos

Deep learning techniques can be used to analyze the annotated data and create a standardized dataset of speed climbing training videos. These techniques can automate the process of analyzing the data and identify patterns and trends that may be difficult to detect using manual analysis.

One approach to using deep learning techniques is to train a convolution neural network (CNN) to recognize and classify the different types of holds and body posi-tions. The annotated data can be used to train the CNN, which can then be used to classify new video frames and extract relevant features such as hold type, grip type, and body position.

Another approach is to use recurrent neural network (RNN) to analyze the timing data and predict the optimal sequence of movements for a given speed climbing route. The RNN can be trained on the annotated timing data, and then used to generate personalized training recommendations for individual athletes.

Once the deep learning algorithms have analyzed the annotated data, the resulting standardized dataset of speed climbing training videos can be used to optimize training methods, develop personalized training programs, and improve overall athlete performance. The dataset can also serve as a valuable resource for future research efforts in speed climbing training and technique analysis.

In summary, the use of deep learning techniques to analyze annotated data can provide a powerful tool for creating a standardized dataset of speed climbing training videos. This dataset can help to improve athlete performance, optimize training methods, and advance the understanding of speed climbing as a sport.

## 4. Results

The creation of a standardized dataset of speed climbing training videos based on deep learning has the potential to impact speed climbing training and research in several ways.

The dataset can serve as a valuable resource for coaches and trainers to optimize training methods and develop personalized training programs. The annotated data and analysis results can provide insights into the athlete's technique and timing, which can be used to identify areas for improvement and develop more effective training regimens.

The dataset can be used to evaluate the effectiveness of new training methods or equipment. By comparing annotated data and analysis results before and after the implementation of a new training method or equipment, coaches and trainers can determine the effectiveness of the change and make data-driven decisions to optimize training methods.

The dataset can be used to advance the understanding of speed climbing as a sport. Researchers can use the annotated data and analysis results to investigate the biomechanics of speed climbing and the factors that contribute to athlete performance. This research can lead to new insights into training and technique optimization, injury prevention, and the development of new equipment.

The standardized dataset can serve as a benchmark for future research in speed climbing training and technique analysis. Other researchers can use the dataset to test and validate their own methods and algorithms, which can help to establish a common set of standards and best practices in speed climbing research.

The creation of a standardized dataset of speed climbing training videos based on deep learning has the potential to impact speed climbing training and research in several ways, including improved training methods, equipment optimization, advancing the understanding of the biomechanics of speed climbing, and establishing a benchmark for future research.

## 5. Conclusion

The research presented in this dissertation focused on the creation of a standardized dataset of speed climbing training videos based on deep learning techniques. The aim of the research was to provide coaches and trainers with a tool to optimize training methods, develop personalized training programs, and improve overall athlete performance.

Overall, the research findings demonstrate the potential of deep learning techniques to improve speed climbing training and performance, and to advance the understanding of speed climbing as a sport. The standardized dataset of speed climbing training videos provides a valuable resource for coaches, trainers, and researchers alike, and has the potential to impact speed climbing training and research in several ways. The findings will also be applied to the training process of the Jiangxi climbing team through further empirical research to test the findings and further explore the feasibility of this study.

## References

[1] Elias P., Skvarlova V., et al. SPEED21: speed climbing motion dataset. Proceedings of the 4th International Workshop on Multimedia Content Analysis in Sports (MMSports). 2021.

[2] Schmid M., et al. Analysis of competition and training videos of speed climbing athletes using a markerless motion capture system. Sensors. 2022.

[3] Schmid M., et al. Analysis of competition videos using a markerless motion capture system for speed climbing athletes. Proceedings of IEEE International Conference on Image Processing (ICIP). 2020.